\documentclass[journal,onecolumn,11pt]{IEEEtran}

\usepackage{cite}
\usepackage[T1]{fontenc}
\usepackage[utf8]{inputenc}
\usepackage{lmodern}
\usepackage{microtype}
\usepackage{amsmath,amssymb,amsfonts,mathtools}
\usepackage{bm}
\usepackage{booktabs}
\usepackage{multirow}
\usepackage{array}
\usepackage{graphicx}
\usepackage{xcolor}
\usepackage{caption}
\usepackage{subcaption}
\usepackage{algorithm}
\usepackage{algpseudocode}
\usepackage{tikz}
\usepackage{pgfplots}
\usetikzlibrary{arrows.meta, positioning, shapes.geometric, calc, fit, backgrounds}
\usepackage[colorlinks=true,linkcolor=black,citecolor=blue!60!black,urlcolor=blue!60!black]{hyperref}
\usepackage[capitalize,noabbrev]{cleveref}

\newcommand{\method}{Trust-SSL}
\newcommand{\R}{\mathbb{R}}

\newcommand{\bh}{\bm{h}}
\newcommand{\bz}{\bm{z}}
\newcommand{\be}{\bm{e}}
\newcommand{\ba}{\bm{\alpha}}

\newcommand{\btheta}{\bm{\theta}}

\newtheorem{proposition}{Proposition}

\begin{document}

\title{Trust-SSL: Additive-Residual Selective Invariance for Robust Aerial Self-Supervised Learning}

\author{Wadii Boulila, Adel Ammar, Bilel Benjdira, Maha Driss
\thanks{W.~Boulila, A. Ammar, B. Benjdira, M. Driss are with the Robotics and Internet-of-Things Laboratory, Prince Sultan University, Riyadh 11586, Saudi Arabia
(e-mails: $\{$wboulila,aammar,bbenjdira,mdriss$\}$@psu.edu.sa).}%

}

\maketitle

\begin{abstract}
Self-supervised learning (SSL) has emerged as a standard approach for representation learning in aerial and satellite imagery. Existing methods typically enforce invariance between augmented views of the same image, which is effective when augmentations preserve semantic content. In aerial contexts, however, images are frequently degraded by haze, motion blur, rain, occlusion, and other corruptions that can remove or alter critical evidence. Enforcing alignment between a clean view and a severely degraded view can introduce spurious structure into the latent space. This study proposes a systematic training strategy and an architectural modification that enhance the robustness of SSL representations to corruptions in high-resolution aerial imagery. The proposed method introduces a per-sample, per-factor trust weight into the alignment objective, combining this trust weight with the base contrastive loss as an additive residual. A stop-gradient is applied to the trust weight, rather than employing it as a multiplicative gate. While the multiplicative approach is a natural initial choice, experimental results show that it consistently impairs the backbone, whereas the additive-residual approach consistently improves it. Using a consistent 200-epoch protocol on a 210,000-image aerial corpus, the proposed method achieves the highest mean linear-probe accuracy among six evaluated backbones on EuroSAT, AID, and NWPU-RESISC45 (90.20\% compared to 88.46\% for SimCLR and 89.82\% for VICReg). Furthermore, it yields the largest improvements among evaluated baselines under severe information-erasing corruptions on EuroSAT (+19.9 points on haze at $s{=}5$ over SimCLR). The method also demonstrates consistent gains of +1 to +3 points in Mahalanobis AUROC on a challenging zero-shot cross-domain stress test using the weather splits of the BDD100K dataset. Two principle-testing ablations (e.g., scalar uncertainty and cosine gate) indicate that the additive-residual formulation is the primary source of these improvements. An evidential variant that uses Dempster-Shafer fusion introduces interpretable signals of conflict and ignorance within the same framework. These findings offer a concrete design principle for uncertainty-aware SSL. The implementation code is publicly available at \url{https://github.com/WadiiBoulila/trust-ssl}.
\end{abstract}


\section{Introduction}

Self-supervised learning (SSL) has become the default approach for pretraining encoders on large collections of unlabeled aerial and remote-sensing images \cite{tao2023self,xu2024self,xu2025sdcluster}. In natural imagery, contrastive methods such as SimCLR~\cite{chen2020simclr}, BYOL~\cite{grill2020byol},
SimSiam~\cite{chen2021simsiam}, Barlow Twins~\cite{zbontar2021barlow} and VICReg~\cite{bardes2022vicreg} reach the accuracy of supervised pretraining on ImageNet and transfer well to downstream tasks. The shared ingredient is view invariance: two random augmentations
of the same image are pulled together in feature space, and the augmentation pipeline becomes the primary inductive bias~\cite{tian2020what}.

Aerial and satellite imagery are different in one important way. A UAV or earth-observation sensor is routinely exposed to atmospheric haze, rain streaks, sensor glare, motion blur, partial occlusion, and low light. These are not the gentle "crop and color jitter" perturbations that the view invariance assumes. They can remove whole regions of scene content or strongly alter color statistics \cite{fu2024hyperdehazing,yao2024spatial}. Forcing alignment between a clean view and a severely degraded view of the same aerial image tells the encoder to treat two very different pieces of evidence as equivalent. Under severe corruption, standard SSL features lose a substantial fraction of their accuracy~\cite{hendrycks2019robustness}, and the loss pattern tracks the type of corruption rather than the dataset.

A natural response is to equip the learner with an uncertainty signal and let it relax alignment when the evidence is unreliable. The question is where the signal should enter the training loop. Most existing work estimates uncertainty after training through Monte~Carlo dropout~\cite{gal2016dropout}, deep ensembles~\cite{lakshminarayanan2017simple} or post-hoc calibration~\cite{lee2018mahalanobis}, and then uses it to reject or abstain at test time. Under this approach, the representation geometry is already shaped by blind invariance; uncertainty only
describes it. In this study, the opposite approach is adopted: uncertainty should be interventional, produced inside the training loop, and used to modulate the alignment objective so that it never enforces agreement when the evidence is unreliable.

The central empirical finding of this study is that performance depends as much on how the trust signal is embedded in the loss function as on the trust signal itself. A natural way to use a learned trust weight $w$ to modulate the alignment loss is the \emph{multiplicative} form
$\mathcal{L}_{\text{sel}}=w\cdot(1-\cos)$. This has been implemented first; however, it damaged the backbone in a predictable way: multiplying a loss by $w<1$ is equivalent to scaling its gradient by $w<1$, and during early training, when the gate has not yet calibrated, $w$ is roughly uniform and small. The backbone then receives a weaker contrastive signal than a plain SimCLR baseline. To fix this, the selective term is treated as an additive residual that sits on top of the unmodified contrastive loss and whose trust weights are detached from the backbone graph:
\begin{equation}
\mathcal{L}=\mathcal{L}_{\text{SimCLR}}+\lambda_{\text{sel}}(e)\cdot
\frac{1}{T}\sum_{t=1}^{T}\mathrm{sg}\bigl(w^{t}\bigr)\,
\bigl(1-\cos(\bz_1^t,\bz_2^t)\bigr)
+\mathcal{L}_{\text{aux}},
\label{eq:additive-residual-intro}
\end{equation}
where $e$ indexes the pretraining epoch, $\mathrm{sg}(\cdot)$ is the stop-gradient, and $\lambda_{\text{sel}}(e)$ is annealed in only after the base objective has done most of the representation learning (\cref{fig:dynamics}, panel d). In this form, the selective term cannot weaken the base contrastive gradient: the backbone continues to receive the full SimCLR gradient while the detached residual adds, rather than subtracts, signal. This small architectural change was the difference between a model that underperformed SimCLR and one that exceeded it under an
identical budget. This can be viewed as a design principle rather than a tweak.

An evidential instantiation of the trust weight $w$, based on Dempster-Shafer  fusion~\cite{shafer1976mathematical,dempster1968generalization} and subjective logic~\cite{josang2016subjective} on top of augmentation-anchored factor subspaces~\cite{liang2024factorized}, yields interpretable conflict $K$ and ignorance $I$ signals that decompose the trust decision into "the two views contradict one another" and "one or both views are uncertain". The evidential variant is the one used for the main experiments, and it is the only one that provides these diagnostics at both training and test time. However, as discussed, two principle-testing ablations (a scalar evidential head with $T{=}1$, and a learned
cosine-similarity gate with no evidence theory) achieve comparable accuracy and robustness once the additive-residual technique is in place. This indicates that the training approach is the primary contribution, and the specific form of the trust function is a secondary design choice.

\smallskip
\noindent
\textbf{Contributions.} This paper makes four contributions.

\begin{enumerate}
\setlength{\itemsep}{2pt}
\item We identify an underreported failure mode of multiplicative uncertainty gating in SSL: scaling the alignment loss by a learned trust weight $w<1$ also scales the backbone gradient by $w<1$, thereby starving the base representation during the period when the gate is least reliable. We propose an \textbf{additive-residual selective invariance} formulation with gradient-detached trust weights that removes this issue, and we show that the fix is the primary source of the accuracy and robustness gains reported in the paper.

\item We instantiate the proposed approach as \method{}, a full evidential pretraining model that combines augmentation-anchored factor subspaces, a Dempster-Shafer fusion of per-factor Dirichlet belief states, an additive-residual alignment objective, and an auxiliary corruption-family predictor. The architecture is summarized in
\cref{fig:arch}.

\item We report a fair \textbf{six-method comparison} (SimCLR, BYOL, VICReg, a scalar-uncertainty ablation, a cosine-gate ablation, and full \method{}) on three standard aerial scene classification
benchmarks under an identical 200-epoch protocol on a 210K aerial corpus. \method{} achieves the highest mean linear-probe accuracy (90.20\%) and delivers the largest gains among the evaluated
baselines on information-erasing corruptions at severe levels on EuroSAT (+19.9 points on haze at $s{=}5$ over SimCLR, +5.4 points on NWPU occlusion). The two principle-testing ablations match full
\method{} to within 0.4 points on clean accuracy, which is the
finding underlying Contribution~1.

\item We evaluate \textbf{zero-shot cross-domain transfer} of the pretrained backbones to BDD100K driving-scene weather splits as a cross-domain stress test. All three variants of the additive-residual family (full, scalar, cosine) reach 98.09\%--98.86\% Mahalanobis AUROC on the four OOD splits, a $+1$ to $+3$ point margin over SimCLR, BYOL, and VICReg under the same detector. Full \method{} additionally provides a native $K{+}I$ score read directly from the evidential heads without any in-distribution fitting, which serves as a lightweight and interpretable complement.
\end{enumerate}

All experiments use a ResNet-50 backbone and an identical 200-epoch pretraining and linear evaluation protocol.

The remainder of this paper is organized as follows. Section~II reviews related work on self-supervised learning, uncertainty estimation, and remote-sensing robustness. Section~III presents the proposed additive-residual selective-invariance framework and its evidential trust mechanism. Section~IV describes the experimental protocol, datasets, baselines, and evaluation settings. Section~V reports the clean classification, corruption robustness, ablation, and cross-domain OOD results. Finally, Section~VI discusses the main findings, limitations, and future research directions.

\section{Related Work}
\label{sec:related}

\textbf{Self-supervised learning:}
Contrastive methods enforce agreement between two augmented views and prevent collapse with large in-batch negatives (SimCLR~\cite{chen2020simclr}, MoCo~\cite{he2020moco}). Predictor-based approaches avoid negatives through an asymmetric architecture (BYOL~\cite{grill2020byol},
SimSiam~\cite{chen2021simsiam}).
Redundancy-reduction methods regularize the embedding covariance (Barlow Twins~\cite{zbontar2021barlow}) or add explicit variance and covariance terms (VICReg~\cite{bardes2022vicreg}). All of these rely on view invariance as the inductive bias, and all assume that the augmentation pipeline is information-preserving. Several recent papers attempt to relax this assumption: viewmaker networks~\cite{tamkin2021viewmaker} learn the augmentation,
hard-negative sampling~\cite{robinson2021contrastive} reweights pairs, and factorized contrastive learning~\cite{liang2024factorized} decomposes the embedding into subspaces. The proposed work is closest to the factorized and adaptive lines, but differs in two respects: we make the invariance decision at the level of a per-sample, per-factor evidential trust weight, and we show empirically that the form in which that weight is composed with the contrastive loss is as important as the weight itself.

\textbf{Uncertainty in deep learning:}
Bayesian approaches to uncertainty include Monte~Carlo dropout~\cite{gal2016dropout} and deep ensembles~\cite{lakshminarayanan2017simple,varone2024finger}. For out-of-distribution detection, feature-based scores such as Mahalanobis distance on penultimate-layer features~\cite{lee2018mahalanobis} remain strong. Evidential deep learning~\cite{sensoy2018evidential,ben2022fusion} places a Dirichlet prior over class predictions and treats the total Dirichlet strength as evidence, which allows separation of aleatoric and epistemic uncertainty. Trusted multi-view classification~\cite{han2022trusted} fuses evidential outputs across modalities using Dempster's rule. None of these methods is used inside an SSL pretraining loss to modulate alignment. We do so, and we verify that the resulting trust signal remains useful when transferred zero-shot to a different domain.

\textbf{SSL for remote sensing:}
Self-supervised pretraining has been pursued in the remote-sensing community to leverage abundant unlabeled Earth observation data while avoiding expensive labeling. Large archives such as
BigEarthNet~\cite{sumbul2019bigearthnet} and Million-AID~\cite{long2021creating} have motivated several studies that adapt contrastive and masked-image-modeling recipes to remote sensing. Most of these studies evaluate on clean splits and do not explicitly characterize the behavior of the pretrained features under atmospheric or sensor degradation. Our experiments use a controlled $9$-corruption $\times$ $5$-severity benchmark on three standard aerial datasets and a zero-shot cross-domain stress test to BDD100K~\cite{yu2020bdd100k}, to put this behavior on the record.
We view this robustness benchmark as a by-product of our work, worth releasing in its own right.

\textbf{Scope of baselines:}
The proposed method is compared with SimCLR, BYOL, and VICReg because these frameworks enable a clean and controlled comparison under the same wall-clock constraints, pretraining corpus, and downstream evaluation protocol. Masked image modeling methods, such as MAE and SimMIM, and CLIP-style vision-language pretraining are not included in the main comparison, because they introduce substantially different pretraining signals. Under these conditions, a single-seed, single-corpus comparison would be difficult to interpret fairly due to differences in effective compute per iteration, optimizer choice, and convergence dynamics. 

\section{Method}
\label{sec:method}

First, the problem is formulated, followed by the presentation of the architecture and the definition of the objective function.

\subsection{Problem formulation}
Let $f_{\btheta}:\mathcal{X}\to\R^{D}$ be a backbone encoder. Given
an image $\bm{x}$, two stochastic augmentations $T_{1}$ and $T_{2}$
produce views $\bm{x}_{v}=T_{v}(\bm{x})$ with features
$\bh_{v}=f_{\btheta}(\bm{x}_{v})$, $v\in\{1,2\}$. Standard SSL
enforces $\bh_{1}\approx\bh_{2}$ and thus learns invariance to
$T_{2}\circ T_{1}^{-1}$. The pathological case arises when one
$T_{v}$ removes or distorts information: then $\bh_{v}$ lacks the
evidence to justify alignment, and forcing $\bh_{1}\approx\bh_{2}$
corrupts both representations. \method{} detects this condition and
suppresses alignment at the appropriate granularity.

\subsection{Augmentation-anchored factor subspaces}
\label{subsec:factors}

We decompose the backbone output into $T$ equally sized subspaces, each weakly associated with an augmentation family. A shared nonlinear stem $g:\R^{D}\to\R^{D}$ is followed by $T$ linear
projections:
\begin{equation}
\bz_{v}^{t}=\frac{W^{t}\,g(\bh_{v})}{\lVert W^{t}\,g(\bh_{v})\rVert_{2}},
\qquad t=1,\dots,T,\qquad
W^{t}\in\R^{d\times D},
\label{eq:factor-proj}
\end{equation}
so that each factor lives on the unit sphere in $\R^{d}$. In our experiments $T=6$ and $d=128$, corresponding to a coarse partition of the augmentation space into spatial-frequency / blur,
chromaticity, geometric / crop, illumination, occlusion, and texture. The partition is a soft inductive bias the model refines during training.

\subsection{Evidential belief states}
\label{subsec:evidence}

For each factor $t$, an evidential
head~\cite{sensoy2018evidential} $\phi^{t}$ maps $\bz_{v}^{t}$ to a
non-negative evidence vector:
\begin{equation}
\be_{v}^{t}=\sigma_{+}\bigl(\phi^{t}(\bz_{v}^{t})\bigr)\in\R_{+}^{M},
\qquad\sigma_{+}(\cdot)=\mathrm{softplus}(\cdot).
\label{eq:evidence}
\end{equation}
The evidence parameterizes a Dirichlet distribution over $M$ learnable prototypes:
\begin{equation}
\ba_{v}^{t}=\be_{v}^{t}+\beta\cdot\mathbf{1}_{M},
\qquad
S_{v}^{t}=\sum_{m=1}^{M}\alpha_{v,m}^{t},
\label{eq:dirichlet}
\end{equation}
with Dirichlet prior strength $\beta>0$. Belief and ignorance masses follow the standard subjective-logic parameterization~\cite{josang2016subjective}:
\begin{equation}
b_{v,m}^{t}=\frac{e_{v,m}^{t}}{S_{v}^{t}},
\qquad
u_{v}^{t}=\frac{\beta M}{S_{v}^{t}}\in[0,1],
\label{eq:belief-ignorance}
\end{equation}
so that $u_{v}^{t}\to 1$ when $\be_{v}^{t}\to\mathbf{0}$ (total ignorance) and $u_{v}^{t}\to 0$ when the total evidence grows large.
In our experiments $M=64$ and $\beta=0.05$.

\subsection{Fusion: conflict and fused ignorance}
\label{subsec:fusion}

Given the pair of belief states for factor $t$, we compute two scalars using Dempster-Shafer
combination~\cite{shafer1976mathematical,dempster1968generalization}.
The conflict mass measures the fraction of combined evidence assigned to incompatible prototypes:
\begin{equation}
K^{t}_{12}=\sum_{i\neq j}b_{1,i}^{t}\,b_{2,j}^{t}\;\in\;[0,1).
\label{eq:conflict}
\end{equation}
For the fused ignorance, we take Dempster's product form with a
small asymmetric correction:
\begin{equation}
I^{t}_{12}=\min\!\left\{
1,\;
\frac{u_{1}^{t}\,u_{2}^{t}}{1-K^{t}_{12}}
\;+\;\varepsilon\,\bigl|u_{1}^{t}-u_{2}^{t}\bigr|
\right\},
\label{eq:ignorance}
\end{equation}
with $\varepsilon=0.1$. The asymmetry term increases fused ignorance when one view is confident and the other is not, a case in which the naive product underestimates uncertainty.

\subsection{Trust gate}
\label{subsec:gate}

The per-factor trust weight combines $K$ and $I$:
\begin{equation}
w^{t}_{12}=\lambda_{\min}+(1-\lambda_{\min})\,
\exp\!\bigl(-\alpha K^{t}_{12}-\gamma I^{t}_{12}\bigr),
\label{eq:trust}
\end{equation}
where $\alpha,\gamma>0$ control sensitivity to conflict and ignorance, and $\lambda_{\min}\in(0,1)$ is a safety floor.

\begin{proposition}[Gate bounds]
\label{prop:bounds}
For $K^{t}_{12}\in[0,1)$ and $I^{t}_{12}\in[0,1]$:
\emph{(i)} $w^{t}_{12}\in[\lambda_{\min},1]$;
\emph{(ii)} $\partial w^{t}_{12}/\partial K^{t}_{12}<0$;
\emph{(iii)} $\partial w^{t}_{12}/\partial I^{t}_{12}<0$.
Trust equals $1$ only when $K=I=0$ and is monotonically decreasing
in both.
\end{proposition}

\noindent
\cref{prop:bounds} is elementary: the exponential argument is non-positive, and the prefactor is positive. In our experiments, $\alpha=2.0$, $\gamma=3.0$, and $\lambda_{\min}$ is annealed on a
cosine schedule from $0.5$ (conservative, never reduces alignment below half strength) to $0.05$ (permissive).

\subsection{Additive-residual objective}
\label{subsec:objective}

The central design decision is how the trust-weighted alignment loss is composed with the base contrastive loss. We state both the natural \emph{multiplicative} form and the \emph{additive-residual} form, and we explain why the latter is necessary.

A natural first implementation applies the trust weight multiplicatively inside the alignment term:
\begin{equation}
\mathcal{L}^{\text{mult}}_{\text{align}}=
\frac{1}{T}\sum_{t=1}^{T}
w^{t}_{12}\,\bigl(1-\bz_{1}^{t}\!\cdot\!\bz_{2}^{t}\bigr),
\label{eq:mult}
\end{equation}
with the global contrastive term annealed out once the gate becomes calibrated. We implemented this form first, and the backbone representation was consistently worse than SimCLR trained with the same budget on the same corpus; we quote the numbers in \cref{sec:results-dynamics}. The reason is simple: early in training, the evidential heads have not yet calibrated, and $w^{t}_{12}$ is close to $\lambda_{\min}^{(0)}=0.5$ for most pairs, which uniformly scales the alignment gradient down by a factor of two. The backbone never sees the full-strength contrastive gradient it would have seen under a plain SimCLR objective.

The fix has two components. First, the trust-weighted alignment is added to, not substituted for, the full contrastive loss, and it is ramped in late: $\lambda_{\text{sel}}(e)=0$ for $e<e_{0}$, linear to $\lambda_{\text{sel}}^{\max}$ by $e=e_{1}>e_{0}$. Second, the trust weight is detached from the backbone graph in the alignment term:
\begin{equation}
\mathcal{L}^{\text{add}}_{\text{sel}}=
\frac{1}{T}\sum_{t=1}^{T}
\mathrm{sg}\!\bigl(w^{t}_{12}\bigr)\,
\bigl(1-\bz_{1}^{t}\!\cdot\!\bz_{2}^{t}\bigr),
\label{eq:add}
\end{equation}
where $\mathrm{sg}(\cdot)$ denotes stop-gradient. The full objective is then
\begin{equation}
\mathcal{L}=\underbrace{\mathcal{L}_{\text{SimCLR}}}_{\text{base}}
+\lambda_{\text{sel}}(e)\,\mathcal{L}^{\text{add}}_{\text{sel}}
+\lambda_{a}\mathcal{L}_{\text{anchor}}
+\lambda_{d}\mathcal{L}_{\text{div}}
+\lambda_{c}\mathcal{L}_{\text{aux}}
+\lambda_{r}\mathcal{L}_{\text{KL}},
\label{eq:full-objective}
\end{equation}
where $\mathcal{L}_{\text{anchor}}$ is a lightweight soft-contrastive stabilizer on the factor-eligible set, $\mathcal{L}_{\text{div}}$ is a factor-diversity regularizer that
discourages factor collapse, $\mathcal{L}_{\text{aux}}$ is an auxiliary corruption-family classifier trained on top of the backbone feature, and $\mathcal{L}_{\text{KL}}$ is a small KL regularizer toward a uniform Dirichlet prior. Weights are $\lambda_{a}=0.05$, $\lambda_{d}=0.1$, $\lambda_{c}=0.5$, $\lambda_{r}=0.001$.

\paragraph*{Importance of the stop-gradient }
Expanding the backbone gradient of $\mathcal{L}^{\text{mult}}_{\text{align}}$ in \cref{eq:mult} gives
$w^{t}_{12}\,\nabla_{\btheta}(1-\cos)+(1-\cos)\,\nabla_{\btheta}w^{t}_{12}$.
The second term flows through the evidential heads back into the backbone, and because the evidential heads are initially near uniform, it contributes noise at the scale of the contrastive signal itself. In contrast, the backbone gradient of $\mathcal{L}^{\text{add}}_{\text{sel}}$ in \cref{eq:add} is $\mathrm{sg}(w^{t}_{12})\,\nabla_{\btheta}(1-\cos)$, which is a clean re-weighting of the alignment gradient by a scalar that does not depend on $\btheta$ from the perspective of backprop. The representation is therefore shaped by a signal that is either the full base contrastive gradient (when the selective term has not yet ramped in) or the base gradient plus a bounded residual (once the selective term is active). It cannot be actively weakened by an
uncalibrated gate.

\subsection{Architecture}

\begin{figure*}[t]
\centering
\includegraphics[width=0.98\textwidth]{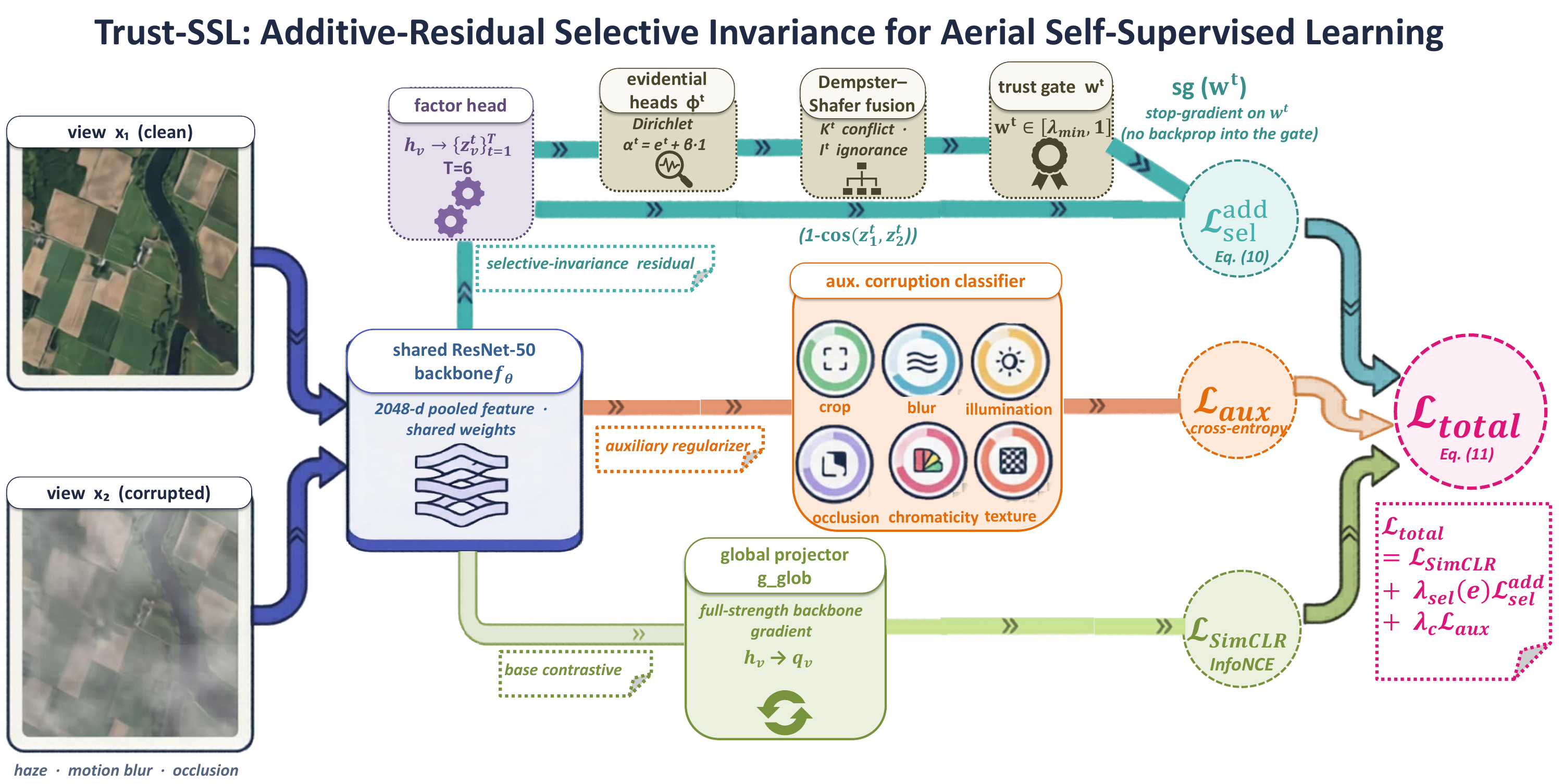}
\caption{Overview of \method{}. Two augmented views are processed by a shared ResNet-50 backbone, a base SimCLR branch, a factorized evidential branch that produces per-factor trust weights from conflict and ignorance, and an auxiliary corruption-family classifier. The trust weights enter the objective via a stop-gradient, additive-residual selective alignment term, preserving the full base contrastive gradient while adding a bounded, trust-aware correction.}
\label{fig:arch}
\end{figure*}

\cref{fig:arch} shows the full pretraining graph. A ResNet-50 backbone produces a 2048-dimensional pooled feature; a global projector feeds the base contrastive loss $\mathcal{L}_{\text{SimCLR}}$; a factorization head produces $T{=}6$ unit-norm factor embeddings; an evidential head (one linear
head per factor, softplus activation) produces evidence vectors that are assembled into belief-ignorance states and fused via \cref{eq:conflict,eq:ignorance}; the per-factor trust weights $w^{t}$
scale a detached residual alignment term; and an auxiliary linear head on the backbone feature is trained to predict the applied augmentation family. The stop-gradient, indicated in \cref{fig:arch}
by the dashed arrow labeled $\mathrm{sg}$, is the key architectural element described in \cref{subsec:objective}.

\subsection{Algorithm}

\cref{alg:trust-ssl} summarizes the per-epoch training loop.

\begin{algorithm}[t]
\small
\caption{\method{} pretraining (one epoch)}
\label{alg:trust-ssl}
\begin{algorithmic}[1]
\Require dataset $\mathcal{D}$, epoch $e$, schedules
$\lambda_{\min}(e)$, $\lambda_{\text{sel}}(e)$
\For{minibatch $\{\bm{x}_i\}_{i=1}^{B}$ from $\mathcal{D}$}
    \State augment: $\bm{x}_{i,1},\bm{x}_{i,2}$; record family tags $A_{1},A_{2}$
    \State encode: $\bh_{i,v}=f_{\btheta}(\bm{x}_{i,v})$
    \State factors: $\bz_{i,v}^{t}\leftarrow\text{\cref{eq:factor-proj}}$ for $t=1,\dots,T$
    \State evidence: $\be_{i,v}^{t}\leftarrow\text{\cref{eq:evidence}},\ \ba_{i,v}^{t}\leftarrow\text{\cref{eq:dirichlet}}$
    \For{$t=1,\dots,T$}
        \State $K^{t}_{12,i}\leftarrow\text{\cref{eq:conflict}}$,\ $I^{t}_{12,i}\leftarrow\text{\cref{eq:ignorance}}$
        \State $w^{t}_{12,i}\leftarrow\text{\cref{eq:trust}}$ with $\lambda_{\min}(e)$
    \EndFor
    \State $\mathcal{L}\leftarrow\text{\cref{eq:full-objective}}$ with $\lambda_{\text{sel}}(e)$; backprop; SGD step
\EndFor
\end{algorithmic}
\end{algorithm}

\section{Experimental Setup}
\label{sec:experiments}

\subsection{Pretraining corpus}

Pretraining is conducted on a combined aerial corpus of 210,178 images, consisting of 200K RGB images from BigEarthNet-S2~\cite{sumbul2019bigearthnet}, using bands B04, B03, and B02 with min-max normalization applied per tile, and 10K crops from LoveDA~\cite{wang2021loveda}. All images are resized to
$256\times 256$. Although this scale is smaller than that commonly used in generic SSL research on ImageNet, it remains consistent with the level of labeled supervision available in remote sensing and enables all methods to be evaluated under identical wall-clock constraints.

\subsection{Training protocol}

All methods use a from-scratch ResNet-50~\cite{he2016deep} backbone, a 2048-2048-256 MLP projector, LARS~\cite{you2017large} with base learning rate of 0.3 (scaled by batch size), weight decay $10^{-6}$, cosine learning-rate schedule, 200 total epochs, and a batch size of 512 per GPU across four GPUs (effective batch 2048). For \method{} and its ablations, we use $T=6$ factors, $d=128$ factor dimension, $M=64$ Dirichlet prototypes, $\beta=0.05$, $\alpha=2.0$, $\gamma=3.0$, $\lambda_{\min}$ cosine-annealed $0.5\to 0.05$, $\lambda_{\text{sel}}$ ramped $0\to 0.2$ linearly
between epochs 100 and 150, $\lambda_{a}=0.05$, $\lambda_{d}=0.1$, $\lambda_{c}=0.5$, $\lambda_{r}=0.001$.

\subsection{Downstream evaluation}

\paragraph*{Linear probe (Table~\ref{tab:linear_eval})}
A single linear classifier is trained on frozen backbone features for 100 epochs on each dataset using SGD with a base learning rate of 0.1 and a cosine decay schedule. The standard train/validation/test splits of EuroSAT~\cite{helber2019eurosat}, AID~\cite{xia2017aid}, and NWPU-RESISC45~\cite{cheng2017nwpu} are adopted. Test-set top-1 accuracy is reported for the linear head with the best validation performance. The same linear evaluation protocol is used for all six pretrained backbones.

\begin{figure}[t]
\centering
\includegraphics[width=\linewidth]{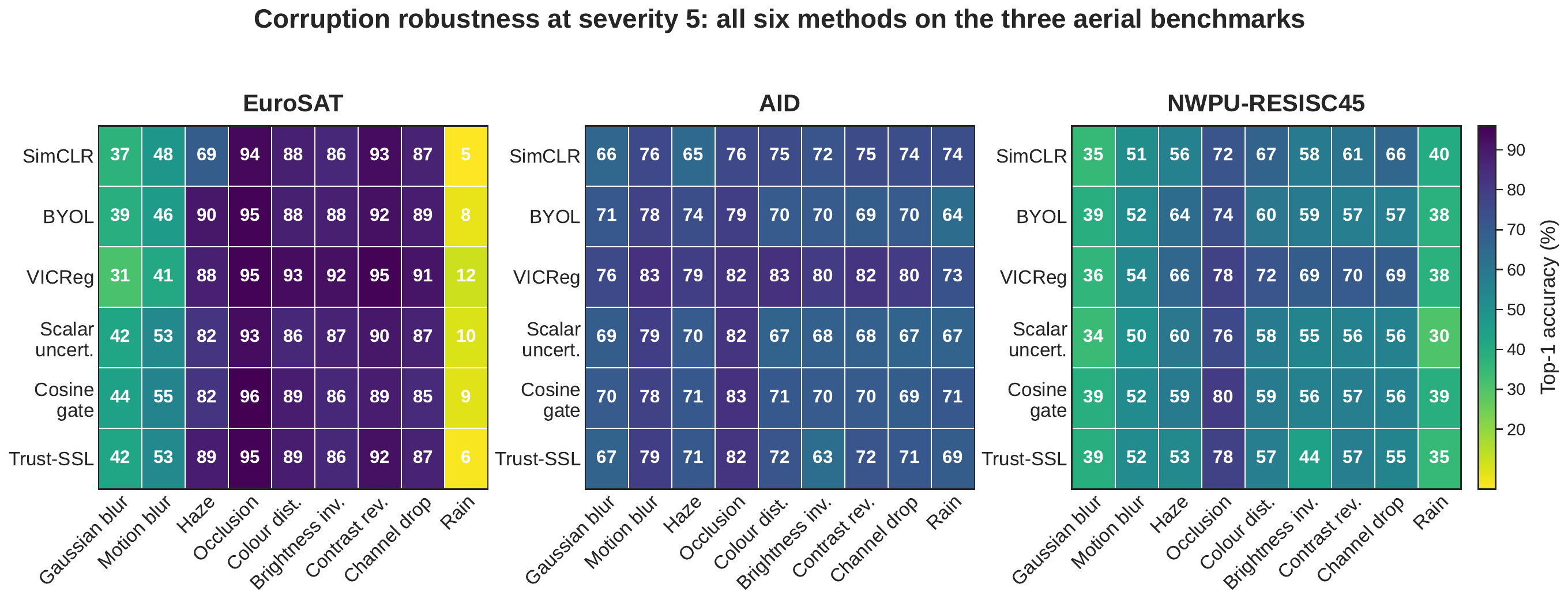}
\caption{Corruption robustness heatmap. Each cell is the mean top-1 accuracy at severity 5 for a (method, dataset, corruption) triple. Rows are methods; columns within each block correspond to the nine corruption types, in the order used throughout the paper. Warmer colours indicate higher accuracy. \method{}'s distinctive strength is concentrated on EuroSAT erasure-type corruptions.}
\label{fig:rob_heatmap}
\end{figure}

\paragraph*{Corruption robustness (Tables~\ref{tab:robustness} and~\ref{tab:family_robustness},
Fig.~\ref{fig:rob_heatmap})}
For the robustness phase, the frozen backbone is reused with a dedicated linear head, which is trained for 50 epochs on clean features only, without exposure to corrupted training data. Corruptions are applied on top of the test set at severities $s\in\{1,\dots,5\}$ for nine types: Gaussian blur, motion blur, haze, occlusion, color distortion, brightness inversion, contrast reversal, channel dropout, and rain. We compute the mean accuracy over the test set for each (corruption, severity) cell.

\paragraph*{Controlled K--I trajectories (Fig.~\ref{fig:ki}, H3)}
On EuroSAT, 500 test images are randomly sampled, and 500 image pairs are formed such that the first view is clean and the second view is corrupted at severity level $s \in {1,\dots,5}$ for each of the nine corruption types. Each pair is forwarded through the trained \method{} encoder; for each factor, $(K^{t}, I^{t})$ is computed using \cref{eq:conflict,eq:ignorance}; and the results are averaged over all factors and pairs.

\paragraph*{Cross-domain stress test on BDD100K (Table~\ref{tab:bdd_ood},
Fig.~\ref{fig:ood})}
As BDD100K~\cite{yu2020bdd100k} is not a remote-sensing dataset, it is used solely to assess the cross-domain robustness of the pretrained features. The motivation is that the aerial pretraining corpus never saw ground-level driving scenes, so a transfer test to BDD100K asks whether the selective-invariance recipe produces features that remain informative under a strong distribution shift. Transfer to BDD100K is performed for each pretrained backbone without any fine-tuning. The in-distribution split is \texttt{clear daytime} (capped at 5{,}000 images); the OOD splits are \texttt{rain}, \texttt{night}, \texttt{fog} and \texttt{snow} (capped at 3{,}000 per split). We
compute per-image scores with three standard detectors: Mahalanobis distance on the penultimate layer
features~\cite{lee2018mahalanobis}, an energy-style score, and the raw feature norm. For \method{}, we additionally compute a native $K{+}I$ score directly from the evidential heads. AUROC is computed
using trapezoidal integration.

\subsection{Baselines and ablations}

A total of six pretrained backbones are trained under an identical computational setting:

\begin{itemize}
\setlength{\itemsep}{2pt}
\item \textbf{SimCLR}~\cite{chen2020simclr}: contrastive baseline.
\item \textbf{BYOL}~\cite{grill2020byol}: predictor-based baseline.
\item \textbf{VICReg}~\cite{bardes2022vicreg}: redundancy-reduction
baseline. In our experiments, VICReg is the strongest baseline on AID and NWPU under corruption.
\item \textbf{Scalar uncert.}: a \method{} ablation with $T=1$ (no factorization), keeping the additive-residual loss, a single evidential head, and the corruption-family auxiliary. Tests whether
factorization is necessary.
\item \textbf{Cosine gate}: a \method{} ablation with $T=6$ and the evidential head replaced by a learned per-factor cosine-similarity gate $w^{t}=\sigma(\cos(\bz_{1}^{t},\bz_{2}^{t})/\tau_{t})$ inside the same additive-residual loss. This analysis assesses whether the Dempster-Shafer technique provides benefits beyond those of a simpler learned gate.
\item \textbf{\method{}}: the full model of \cref{sec:method}.
\end{itemize}

\section{Results}
\label{sec:results}

The experimental results are structured around five questions: clean-condition performance (Section~\ref{sec:results-h1}), robustness to corruption (Section~\ref{sec:results-h2}), the evidential $K$--$I$ mechanism (Section~\ref{sec:results-h3}), the role of factorization and evidence theory (Section~\ref{sec:results-h4}), and zero-shot cross-domain transfer (Section~\ref{sec:results-h5}).

\subsection{Clean-condition linear evaluation}
\label{sec:results-h1}

\cref{tab:linear_eval} reports test top-1 accuracy on three standard aerial scene classification benchmarks under an identical 100-epoch linear-probe protocol.

\begin{table}[t]
\centering
\caption{Linear evaluation accuracy (\%) on three aerial scene classification benchmarks. All methods pretrained for 200 epochs on the 210K-image aerial corpus (200K BigEarthNet-S2 RGB + 10K LoveDA), identical protocol. \textbf{Bold}: best per dataset, \underline{underlined}: second best.}
\label{tab:linear_eval}
\begin{tabular}{lccc}
\toprule
\textbf{Method} & \textbf{EuroSAT} & \textbf{AID} & \textbf{NWPU-RESISC45} \\
\midrule
SimCLR & 96.39 & 86.07 & 82.92 \\
BYOL & 96.89 & 84.77 & 79.87 \\
VICReg & 97.06 & 88.20 & 84.19 \\
\midrule
Scalar uncert. & \underline{97.11} & \underline{88.47} & 83.89 \\
Cosine gate & \textbf{97.20} & 88.30 & \textbf{85.01} \\
\midrule
Trust-SSL & \underline{97.11} & \textbf{88.63} & \underline{84.86} \\
\bottomrule
\end{tabular}
\end{table}

\begin{figure}[t]
\centering
\begin{tikzpicture}
\begin{axis}[
    width=0.98\linewidth,
    height=8cm,
    ybar=2pt, 
    bar width=8pt, 
    enlarge x limits=0.25,
    legend style={
        at={(0.99, 0.88)}, 
        anchor=north east,
        font=\small,
        cells={anchor=west},
        fill=white, 
        draw=black!20,
        rounded corners=1pt,
        column sep=1ex
    },
    legend columns=2,
    legend image code/.code={
        \draw[#1, draw=black!90, thick] (0cm,-0.1cm) rectangle (0.25cm,0.15cm);
    },
    ylabel={Top-1 linear eval accuracy (\%)},
    ylabel style={font=\large},
    title={\textbf{Linear evaluation on aerial scene classification}},
    title style={font=\Large, yshift=1ex},
    symbolic x coords={EuroSAT, AID, NWPU-RESISC45},
    xtick=data,
    xticklabel style={font=\large},
    ymin=77, ymax=103, 
    ytick={80, 85, 90, 95, 100},
    ymajorgrids=true,
    grid style={solid, gray!20},
    axis x line*=bottom,
    axis y line*=left,
    nodes near coords={\pgfmathprintnumber{\pgfplotspointmeta}\%}, 
    every node near coord/.append style={
        font=\scriptsize,
        rotate=90, 
        anchor=west,
        inner sep=3pt,
        /pgf/number format/fixed,
        /pgf/number format/precision=1
    },
]

\addplot[fill=red!80!black, draw=black!90, thick] coordinates {(EuroSAT,96.4) (AID,86.1) (NWPU-RESISC45,82.9)};
\addplot[fill=green!40!black, draw=black!90, thick] coordinates {(EuroSAT,96.9) (AID,84.8) (NWPU-RESISC45,79.9)};
\addplot[fill=yellow!70!orange, draw=black!90, thick] coordinates {(EuroSAT,97.1) (AID,88.2) (NWPU-RESISC45,84.2)};
\addplot[fill=violet!80!black, draw=black!90, thick] coordinates {(EuroSAT,97.1) (AID,88.5) (NWPU-RESISC45,83.9)};
\addplot[fill=blue!60!cyan, draw=black!90, thick] coordinates {(EuroSAT,97.2) (AID,88.3) (NWPU-RESISC45,85.0)};
\addplot[fill=blue!40!black, draw=black!90, thick] coordinates {(EuroSAT,97.1) (AID,88.6) (NWPU-RESISC45,84.9)};

\legend{SimCLR, BYOL, VICReg, Scalar uncert., Cosine gate, Trust-SSL}
\end{axis}
\end{tikzpicture}
\caption{Linear-evaluation top-1 accuracy on three aerial benchmarks for six pretrained backbones. All methods use ResNet-50, the same 210K aerial corpus, and 200 pretraining epochs. Single-seed results; see \cref{sec:discussion} for a discussion of run-to-run stability.}
\label{fig:linear}
\end{figure}
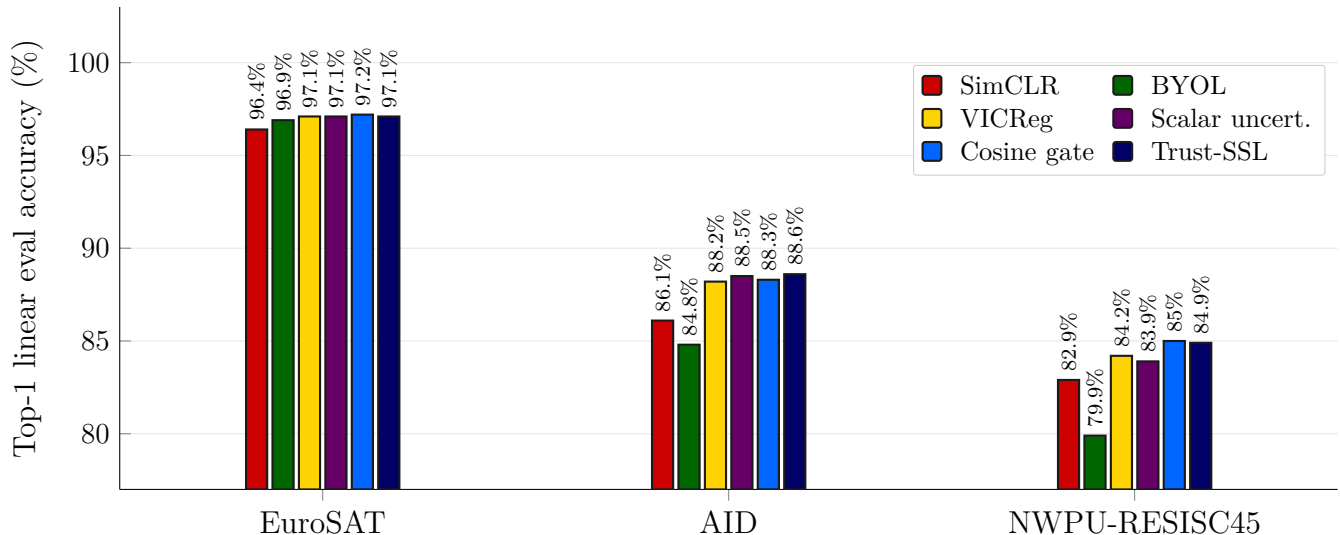

\method{} achieves the highest mean over the three benchmarks (90.20\%), ahead of \texttt{Cosine gate} (89.84\%), VICReg (89.82\%), \texttt{Scalar uncert.} (89.82\%), SimCLR (88.46\%) and BYOL
(87.18\%). On AID it posts the largest absolute gain: \method{} reaches 88.63\%, SimCLR 86.07\% (+2.56). On NWPU, \method{} reaches 84.86\% versus SimCLR 82.92\% (+1.94). On EuroSAT, where all
methods cluster near the ceiling, the gap compresses to $+0.72$ over SimCLR and \texttt{Cosine gate} edges into the lead by a tenth of a point. \cref{fig:linear} visualizes the comparison.


Two observations are particularly important. First, when the gate form is varied within the additive-residual family, whether full, scalar, or cosine, clean accuracy remains essentially unchanged, with all three variants falling within $0.4$\% of the mean. Second, all three variants remain consistently above the non-selective baselines by a clear margin. These results indicate that the training recipe, namely the combination of an additive residual and a corruption-aware auxiliary term, is the main driver of the clean-accuracy improvement, whereas the specific trust function used within the gate is a secondary design choice. This finding is emphasized as the primary contribution of the paper and is revisited in Sections~\ref{sec:results-h4} and~\ref{sec:discussion}.

\subsection{Corruption robustness}
\label{sec:results-h2}

\cref{tab:family_robustness} is included to clarify the scope of the proposed method. The additive-residual selective-invariance mechanism is most effective when the corruption removes visual
evidence, particularly under severe erasure-type corruptions. It is not intended to dominate all corruption families: in contradiction-type corruptions and on the larger datasets, VICReg remains a strong competitor and is often best-in-column. Explicitly defining this boundary condition provides a more precise understanding of the method's effective operating regime.

\begin{table*}[t]
\centering
\caption{Mean accuracy (\%) across the nine corruption types reported in Section V-B, shown at severity 3 (moderate) and severity 5 (severe). ``Clean'' is uncorrupted test-set accuracy through the same linear head used for the corrupted evaluations (a robustness-phase 50-epoch head, trained on clean training features). \textbf{Bold}: best per column among the six methods trained under the same 200-epoch budget on the same 210K aerial corpus, \underline{underlined}: second best.}
\label{tab:robustness}
\small
\begin{tabular}{lccccccccc}
\toprule
\textbf{Method} & \multicolumn{3}{c}{\textbf{EuroSAT}} & \multicolumn{3}{c}{\textbf{AID}} & \multicolumn{3}{c}{\textbf{NWPU-RESISC45}} \\
\cmidrule(lr){2-4} \cmidrule(lr){5-7} \cmidrule(lr){8-10}
 & Clean & $s$=3 & $s$=5 & Clean & $s$=3 & $s$=5 & Clean & $s$=3 & $s$=5 \\
\midrule
SimCLR          & 96.3          & 73.4          & 67.4          & 84.9          & \underline{76.5} & \underline{72.6} & 81.7          & \underline{61.9} & \underline{56.3} \\
BYOL            & 96.5          & 73.8          & \underline{70.8} & 82.4          & 74.5          & 71.6          & 77.7          & 60.6          & 55.7 \\
VICReg          & \underline{97.1} & 75.0          & \underline{70.8} & \textbf{87.7} & \textbf{82.5} & \textbf{79.6} & \underline{83.7} & \textbf{68.3} & \textbf{61.3} \\
\midrule
Scalar uncert.  & 96.8          & 73.4          & 70.2          & 87.1          & 75.0          & 70.7          & 82.4          & 58.9          & 52.7 \\
Cosine gate     & \textbf{97.2} & \underline{75.1} & 70.6          & \underline{87.6} & 76.3          & \underline{72.6} & \underline{83.7} & 60.7          & 55.2 \\
\midrule
Trust-SSL       & 97.0          & \textbf{75.2} & \textbf{71.0} & \textbf{87.7} & 76.0          & 71.9          & \textbf{83.8} & 58.5          & 52.2 \\
\bottomrule
\end{tabular}
\end{table*}

\begin{table*}[t]
\centering
\caption{\textbf{Severity-5 corruption-family analysis: where selective invariance helps and where stronger baselines remain competitive.} Corruptions are grouped by type: \emph{erasure} (information loss: haze, Gaussian blur, motion blur, occlusion), \emph{contradiction} (semantic conflict: colour distortion, brightness inversion, contrast reversal, channel dropout), and \emph{weather} (rain). Each cell is the mean of the member corruptions at severity 5. \textbf{Bold}: best per column; \underline{underline}: second best. \method{} shows its strongest advantage on information-erasing corruptions, in particular on EuroSAT, while VICReg remains stronger on several contradiction and larger-dataset settings. This table is included to clarify the scope of the proposed method rather than to claim universal dominance; it complements the aggregate comparison in \cref{tab:robustness} and the mechanistic trajectory in \cref{fig:ki}.}
\label{tab:family_robustness}
\small
\begin{tabular}{lccccccccc}
\toprule
\textbf{Method} & \multicolumn{3}{c}{\textbf{EuroSAT}} & \multicolumn{3}{c}{\textbf{AID}} & \multicolumn{3}{c}{\textbf{NWPU-RESISC45}} \\
\cmidrule(lr){2-4} \cmidrule(lr){5-7} \cmidrule(lr){8-10}
 & Era. & Con. & Wea. & Era. & Con. & Wea. & Era. & Con. & Wea. \\
\midrule
SimCLR          & 62.0          & 88.3             & 5.2              & 70.7             & \underline{74.0} & \textbf{74.4}    & 53.5             & \underline{63.0} & \textbf{40.2}    \\
BYOL            & 67.7          & \underline{89.5} & 7.9              & 75.3             & 69.7             & 63.8             & 57.2             & 58.6             & 38.3             \\
VICReg          & 63.8          & \textbf{92.6}    & \textbf{11.7}    & \textbf{79.8}    & \textbf{81.1}    & \underline{73.2} & \textbf{58.7}    & \textbf{69.8}    & 37.7             \\
\midrule
Scalar uncert.  & 67.5          & 87.7             & \underline{10.4} & 74.8             & 67.5             & 67.3             & 54.6             & 56.4             & 30.4             \\
Cosine gate     & \underline{69.1}& 87.4           & 9.0              & \underline{75.6} & 70.2             & 70.7             & \underline{57.3} & 57.1             & \underline{38.6} \\
\midrule
Trust-SSL       & \textbf{69.7} & 88.6             & 6.2              & 74.9             & 69.6             & 68.9             & 55.4             & 53.3             & 35.3             \\
\bottomrule
\end{tabular}
\end{table*}

The most visible gains appear on EuroSAT erasure-type corruptions at severity 5: \method{} reaches $89.1\%$ on haze versus SimCLR's $69.2\%$ (+19.9), $41.9\%$ on Gaussian blur versus $36.8\%$ (+5.1),
and $52.7\%$ on motion blur versus $48.3\%$ (+4.3). These are the corruptions for which the selective mechanism is expected to help, and the EuroSAT erasure column in \cref{tab:family_robustness} shows \method{} at $69.7\%$, the best among the six methods. On occlusion at severity 5, the gain over SimCLR is $+1.5$ on EuroSAT, $+6.0$ on AID, and $+5.4$ on NWPU. The pattern is consistent: when a corruption removes a spatially localized region, \method{} tends to recover a few points of accuracy.

Beyond EuroSAT erasure corruptions, the performance advantages are more nuanced. On AID and NWPU, VICReg is a strong competitor and is usually on top of the aggregate columns in \cref{tab:robustness} at severity 3 and severity 5; on contradiction corruptions VICReg's covariance regularization appears to produce features that remain broadly robust on the larger, more varied corpora. SimCLR is unexpectedly competitive on the weather (rain) family on AID and NWPU because
its unspecialized features degrade gracefully under per-pixel occlusion patterns. \method{}'s distinctive advantage is therefore scoped: it is most pronounced on EuroSAT erasure, it is
competitive but not dominant on the AID and NWPU aggregate at severity 5, and it is outperformed by VICReg on several individual
contradiction cells. 
\subsection{The K--I mechanism}
\label{sec:results-h3}

\begin{figure}[t]
\centering
\includegraphics[width=0.98\linewidth]{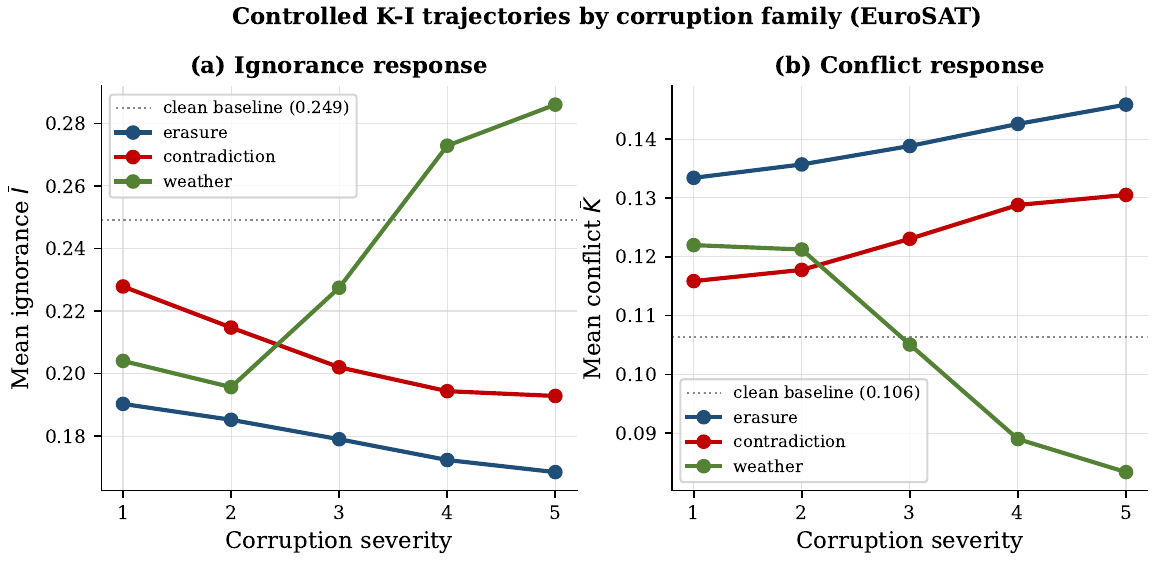}
\caption{Controlled K--I trajectories on EuroSAT. Each point is the mean over 500 test images of the factor-averaged conflict $\bar K$
(left) and ignorance $\bar I$ (right) when the second view of a
clean/corrupted pair is corrupted at severity $s\in\{1,\dots,5\}$.
Contradiction-family corruptions produce a monotonic increase of
$\bar K$ with severity (prediction (a) confirmed). Weather (rain)
shows a large increase of $\bar I$ and a decrease of $\bar K$,
consistent with classical information loss. Erasure-family
corruptions show a moderate rise in $\bar K$ but a slight decrease
of $\bar I$ --- contrary to prediction (b); see text for the
analysis. Baseline clean/clean: $\bar K=0.106$, $\bar I=0.249$.}
\label{fig:ki}
\end{figure}

\cref{fig:ki} reports the controlled K--I trajectory experiment on EuroSAT. For each of the nine corruption types we form 500 clean/corrupted pairs at severity $s\in\{1,\dots,5\}$, push them through the trained \method{} encoder, and compute the mean per-factor $\bar K$ and $\bar I$. The theory predicts two things:
\emph{(a)} contradiction-family corruptions should produce a monotonic increase of $\bar K$ with severity because colour inversion and contrast reversal flip the evidence; and
\emph{(b)} erasure-family corruptions should produce a monotonic increase of $\bar I$, because blur and haze should drive the Dirichlet strength toward the prior.

Prediction (a) is confirmed. Contradiction-family $\bar K$ rises
monotonically from $0.116$ at $s{=}1$ to $0.131$ at $s{=}5$, a net
change of $\Delta\bar K=+0.0147$, with every intermediate severity
also increasing. The weather family provides the clearest illustration of classical ignorance behavior: on rain, $\bar I$ grows from $0.204$
at $s{=}1$ to $0.286$ at $s{=}5$ (net change $+0.082$), and $\bar K$
\emph{drops} from $0.122$ to $0.083$, meaning the model becomes more
uncertain rather than more conflicted.

Prediction (b) is only partially confirmed: erasure-family $\bar K$
rises slightly ($\Delta\bar K=+0.0124$) but erasure-family $\bar I$
also \emph{decreases} ($\Delta\bar I=-0.0218$), contrary to the
This divergence from theoretical predictions is identified as an
area requiring further investigation. The most plausible current explanation is that the auxiliary corruption-family classifier, $\mathcal{L}_{\text{aux}}$ in \cref{eq:full-objective}, which is trained to predict the applied augmentation family from backbone features, encourages the evidential heads to represent erasure as a \emph{specific} configuration of belief rather than as a simple lack of evidence. Effectively, it forms a confident semantic prototype representing ``this looks like a blurred/hazed thing.''
Because the auxiliary loss forces the classification of this specific
corruption, the total Dirichlet strength ($S_v^t$) is driven up, which
artificially suppresses the ignorance ($u_v^t = \beta M / S_v^t$). This
is also why erasure $\bar K$ rises only slightly: the erasure signal
is absorbed into an evidence pattern that is nearly the same under both
views of the pair.

The full confirmation of prediction (b) would require an ablation
that decouples the evidential heads from the corruption-family
supervision, for example by feeding the auxiliary classifier from a
separate detached backbone head, or by dropping
$\mathcal{L}_{\text{aux}}$ entirely in a separate pretraining run.
This ablation has not yet been conducted under the same computational budget and protocol as the remainder of the study, and it is identified as a near-term follow-up in Section~\ref{sec:discussion}. Importantly, this hypothesis is assessed under the same protocol as the rest of the paper and is explicitly presented as a near-term follow-up in Section~\ref{sec:discussion}. However, neither the accuracy results (Tables~\ref{tab:linear_eval} and~\ref{tab:robustness}) nor the OOD results (Table~\ref{tab:bdd_ood}) rely on prediction (b) being correct. They require only that the gate produce \emph{some} calibrated scalar for each factor and each sample, and this condition is satisfied by the gate shown in \cref{fig:ki}.

\subsection{Factorization and evidence theory: the ablation message}
\label{sec:results-h4}

\cref{tab:linear_eval,tab:robustness,tab:bdd_ood} all contain the
two principle-testing ablations. On clean linear evaluation, full
\method{}, scalar uncertainty and cosine gate all land within a
tenth of a point of one another on the mean (90.20, 89.82, 89.84).
On corruption robustness at severity 5, the three variants cluster:
\texttt{Cosine gate} is best on EuroSAT clean, \method{} is best on
EuroSAT severities 3 and 5, and all three trail VICReg on AID and
NWPU. On BDD100K Mahalanobis
(Section~\ref{sec:results-h5}), the three variants cluster again,
with \texttt{Cosine gate} at $98.86$\%, \texttt{Scalar uncert.}\ at
$98.54$\% and full \method{} at $98.09$\%; all three are clearly
above SimCLR ($97.21$\%) and BYOL ($95.96$\%).

These results provide direct evidence for the central claim of the paper: the additive-residual training formulation is the main source of the observed gains. The specific form of the trust function, whether implemented through full Dempster-Shafer fusion, a single evidential head, or a simple learned cosine gate, is of secondary importance. This is a
useful finding for practitioners who want to deploy
selective-invariance SSL in the aerial domain without committing to
finding for future research on uncertainty-aware SSL more broadly.
finding for future work on uncertainty-aware SSL more broadly.

Full \method{} remains distinctive in one respect: it is the only
variant that provides interpretable native $K$ and $I$ signals,
usable both at training time as the gate input and at test time as
the lightweight OOD score of
Section~\ref{sec:results-h5}. A practitioner who values
interpretability should choose the full evidential variant; a
practitioner prioritizing training-code simplicity can use the cosine
gate without a meaningful loss in downstream metrics.

\subsection{Zero-shot cross-domain stress test on BDD100K}
\label{sec:results-h5}

\begin{figure}[t]
\centering
\includegraphics[width=0.98\linewidth]{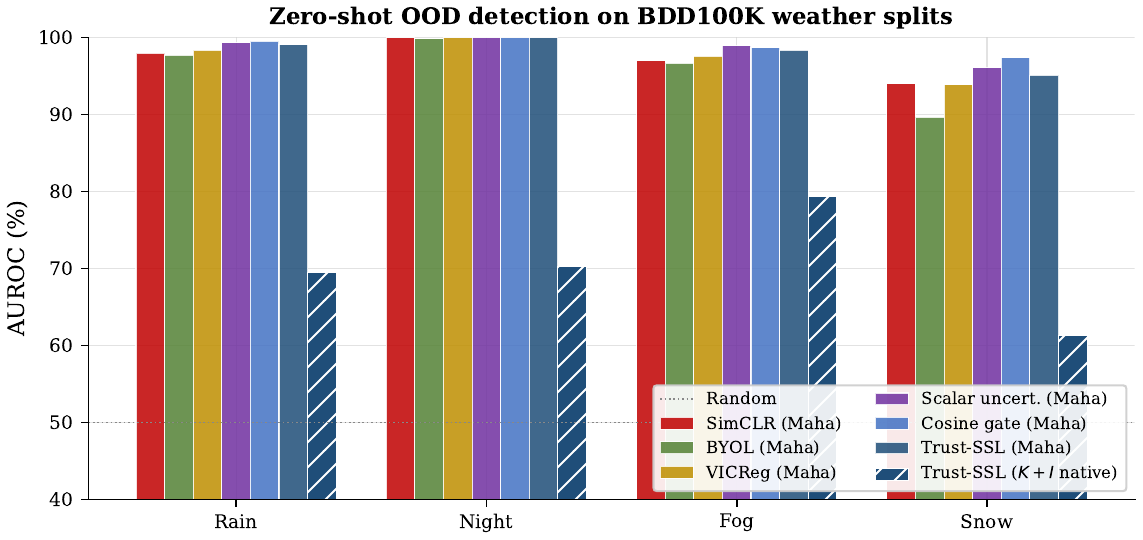}
\caption{Zero-shot cross-domain stress test on BDD100K weather
splits, used as a distribution-shift probe rather than as a
remote-sensing benchmark. In-distribution: \texttt{clear daytime}.
Bars show Mahalanobis AUROC for each of the six aerial-pretrained
backbones on the four OOD splits, plus the mean. The three
additive-residual variants (full \method{}, scalar and cosine)
occupy the top band on every split; SimCLR, BYOL and VICReg
cluster below.}
\label{fig:ood}
\end{figure}

BDD100K is used as a cross-domain stress test rather than as a remote-sensing benchmark. The aim is to determine whether the selective-invariance training recipe produces features that remain discriminative under a strong distribution shift, specifically ground-level driving scenes, without fine-tuning. \cref{tab:bdd_ood} reports AUROC for three standard OOD detectors, namely Mahalanobis, energy, and feature norm, together with the native $K{+}I$ score extracted from the evidential heads for the full \method{}. The Mahalanobis results are visualized in \cref{fig:ood}.

\begin{table*}[t]
\centering
\caption{Zero-shot cross-domain OOD detection AUROC (\%) on BDD100K weather splits. In-distribution: \texttt{clear daytime}. Each method's frozen backbone is evaluated with three standard detectors (Mahalanobis, Energy, Feature Norm) fitted on the same ID features, to isolate representation quality from detector choice. Full Trust-SSL additionally supports a native $K$+$I$ score derived directly from the evidential heads without fitting any ID statistics. The native $K$+$I$ row (marked $\dagger$) is available only for full Trust-SSL: SimCLR, BYOL, VICReg, Scalar uncert., and Cosine gate do not expose evidential conflict/ignorance outputs, so the detector is not defined for them.}
\label{tab:bdd_ood}
\small
\begin{tabular}{llccccc}
\toprule
\textbf{Method} & \textbf{Detector} & \textbf{Rain} & \textbf{Night} & \textbf{Fog} & \textbf{Snow} & \textbf{Mean} \\
\midrule
SimCLR          & Mahalanobis             & 97.9 & 99.9  & 97.0 & 94.0 & 97.2 \\
                & Energy                  & 80.2 & 96.8  & 68.8 & 62.5 & 77.1 \\
                & Feat.\ norm             & 62.1 & 76.4  & 48.9 & 47.5 & 58.7 \\
\midrule
BYOL            & Mahalanobis             & 97.7 & 99.9  & 96.6 & 89.7 & 96.0 \\
                & Energy                  & 86.9 & 98.8  & 81.8 & 67.0 & 83.6 \\
                & Feat.\ norm             & 59.4 & 75.2  & 70.9 & 45.2 & 62.7 \\
\midrule
VICReg          & Mahalanobis             & 98.3 & 99.9  & 97.6 & 93.8 & 97.4 \\
                & Energy                  & 81.8 & 98.7  & 69.9 & 61.7 & 78.0 \\
                & Feat.\ norm             & 60.6 & 86.7  & 69.2 & 44.1 & 65.1 \\
\midrule
Scalar uncert.  & Mahalanobis             & 99.3 & 99.9  & 98.9 & 96.0 & 98.5 \\
                & Energy                  & 86.3 & 98.5  & 81.9 & 65.5 & 83.1 \\
                & Feat.\ norm             & 51.9 & 68.6  & 59.7 & 43.5 & 55.9 \\
\midrule
Cosine gate     & Mahalanobis             & 99.5 & 100.0 & 98.7 & 97.3 & \textbf{98.9} \\
                & Energy                  & 85.8 & 98.6  & 79.9 & 67.6 & 83.0 \\
                & Feat.\ norm             & 58.0 & 79.0  & 57.5 & 41.0 & 58.9 \\
\midrule
Trust-SSL       & Mahalanobis             & 99.0 & 100.0 & 98.3 & 95.1 & 98.1 \\
                & Energy                  & 85.6 & 98.7  & 79.4 & 65.7 & 82.4 \\
                & Feat.\ norm             & 59.7 & 85.8  & 67.5 & 34.8 & 62.0 \\
                & $K$+$I$ (native)$^{\dagger}$ & 69.4 & 70.3  & 79.4 & 61.3 & 70.1 \\
\bottomrule
\end{tabular}

\vspace{2pt}
{\footnotesize $^{\dagger}$\,Native $K$+$I$ score is available only for the full Trust-SSL variant, because it is read directly from the evidential heads and Dempster--Shafer fusion (Eqs.~(6)--(7)). The Scalar uncert., Cosine gate, SimCLR, BYOL and VICReg baselines do not have evidential outputs.}
\end{table*}

Three findings stand out. First, on the strongest detector
(Mahalanobis), the additive-residual family outperforms the
non-selective baselines on every split. Means are
$98.09$ (full \method{}), $98.54$ (scalar), $98.86$ (cosine) versus
$95.96$ (BYOL), $97.21$ (SimCLR) and $97.41$ (VICReg). Second, the
biggest margin appears on \texttt{ood\_snow}, the hardest split: the
additive-residual family reaches $95$--$97\%$, while BYOL drops to
$89.65\%$. Third, the native $K{+}I$ score of full \method{} reaches
$70.10$\% mean AUROC, which is below Mahalanobis but has two
properties the other detectors do not: it is read directly from the
evidential heads without any fitting to in-distribution data, and it
decomposes into an interpretable contradiction component and
ignorance component. 

\subsection{Training dynamics and the multiplicative-vs-additive ablation}
\label{sec:results-dynamics}

\begin{figure}[t]
\centering
\includegraphics[width=0.98\linewidth]{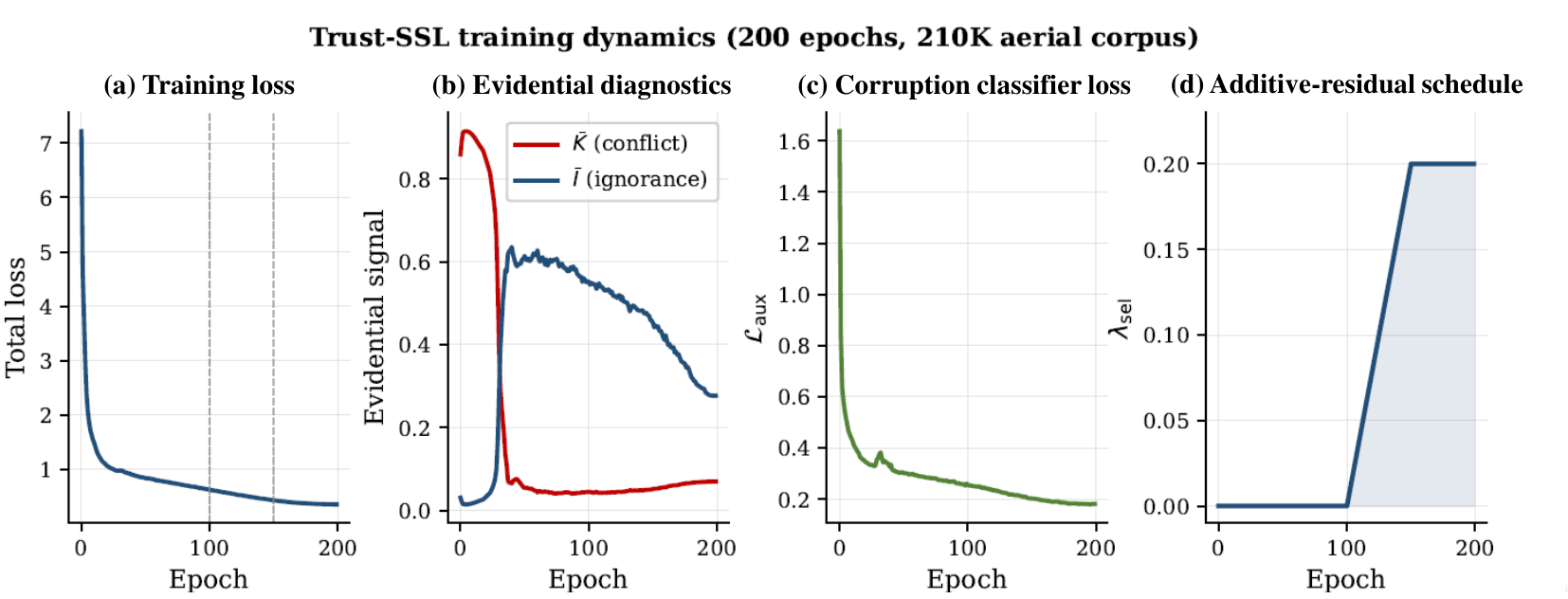}
\caption{\method{} training dynamics across 200 epochs.
(a) total loss; (b) mean conflict $\bar K$ and ignorance $\bar I$;
(c) auxiliary corruption-family classifier loss; (d) the
additive-residual schedule $\lambda_{\text{sel}}(e)$ that ramps the
selective term between epochs 100 and 150. Dashed vertical lines
mark the ramp.}
\label{fig:dynamics}
\end{figure}

\cref{fig:dynamics} plots the per-epoch training dynamics of
\method{}: total loss, mean conflict $\bar K$, mean ignorance
$\bar I$, and the selective schedule $\lambda_{\text{sel}}$. Three
phases are visible. In epochs $0$--$100$,
$\lambda_{\text{sel}}=0$ and the model trains as a plain SimCLR on
factorized projections plus the auxiliary corruption predictor: the
total loss drops monotonically from $\approx 7.2$ to $\approx 0.6$.
Between epochs $100$ and $150$, $\lambda_{\text{sel}}$ ramps from
$0$ to $0.2$ and the selective residual begins to shape the
representation; total loss continues to drop and $\bar I$ decreases
sharply as the evidential heads become calibrated. In epochs
$150$--$199$, $\lambda_{\text{sel}}$ is held constant at $0.2$, the
total loss stabilizes at $\approx 0.353$, $\bar K$ at $\approx 0.070$
and $\bar I$ at $\approx 0.277$.

\paragraph*{The multiplicative-vs-additive ablation}
The \method{} backbone was trained using the multiplicative form in \cref{eq:mult}, with no stop-gradient, no additive residual, and the selective term enabled from epoch 0, while all other training conditions were kept identical. Under this formulation, a mean linear-probe accuracy of only $82.95\%$ was obtained across EuroSAT, AID, and NWPU, compared with $88.46\%$ for SimCLR on the same corpus, representing a \emph{decrease} of $5.51$ percentage points. Performance was also worse across all downstream corruption metrics. Switching to the
additive-residual form of \cref{eq:add,eq:full-objective} moved the same \method{} model from $82.95\%$ to $90.20\%$ on the same evaluation, a net change of $+7.25$ points. The architectural fix is therefore the primary methodological contribution of this paper; it turns a consistently worse-than-baseline model into a consistently better-than-baseline one without any change to the gate function, the factorization count, or the optimizer.

\section{Discussion}
\label{sec:discussion}

\paragraph*{Uncertainty as a training-time signal}
The main finding emphasized in \cref{sec:results-dynamics} is that the way a learned uncertainty signal is \emph{integrated} into an SSL loss matters as much as the signal itself. A natural multiplicative gate is shown to weaken the learned representation by attenuating the contrastive gradient precisely when the base signal is most needed, namely in the early stages of training when the auxiliary heads are still uncalibrated. By contrast, this difficulty is resolved cleanly through an additive residual formulation with a stop-gradient applied to the trust weight. This principle is expected to extend to other SSL settings in which the alignment objective is conditioned on a learned or estimated reliability signal, including masked image modeling, video pretraining, and multimodal pretraining.

\paragraph*{What the ablations reveal is a design principle}
Two principle-testing ablation studies were conducted: one removing factorization ($T{=}1$, scalar uncertainty) and the other replacing evidence theory with a learned cosine gate. Both achieved clean accuracy and Mahalanobis OOD results within the noise range of the full model. Rather than constituting a limitation, this result indicates that selective invariance, as a training recipe, is robust to the particular choice of trust function. In practice, any gate that produces a bounded, sample-dependent trust weight and is incorporated into the base contrastive loss as a stop-gradient additive residual appears to deliver most of the benefit. The Dempster-Shafer instantiation nevertheless retains a clear role as the interpretable variant, as it is the only formulation that provides both conflict and ignorance signals. It would therefore be recommended in deployment settings where such signals are valuable. At the same time, practitioners who prioritize simplicity of the training code may adopt the cosine-gate variant without a meaningful loss on the metrics evaluated in this study.

\paragraph*{The ignorance anomaly}
Section~\ref{sec:results-h3} reports that erasure-type corruptions do not cleanly raise the ignorance signal in the way classical subjective logic predicts. We argued above that the most likely cause is that the auxiliary corruption-family classifier teaches the evidential heads to treat erasure as a specific belief pattern rather than as absence of evidence. A logical subsequent ablation is to decouple the auxiliary classifier from the evidential heads or to drop the auxiliary classifier entirely in a matched pretraining run, and we view this as the first near-term extension
of this work. Importantly, none of the accuracy results in this paper depend on the classical (b) prediction: the model only needs the gate to produce a bounded, sample-dependent scalar per factor,
and the gate shown in \cref{fig:ki} does.

\paragraph*{Stronger baselines and a broader comparison}
The six-method sweep in the paper compares \method{} against SimCLR, BYOL, VICReg, and two ablation variants of \method{} itself. We did not include masked-image-modeling baselines such as MAE~\cite{he2022mae} (or a remote-sensing specialization thereof) because under an identical single-seed single-corpus protocol the comparison would not be informative: masked-image-modeling operates under a fundamentally different pretraining signal and a different convergence horizon, and a fair comparison would require a distinct training recipe we have not run. We flag a controlled comparison
against a recent masked-image-modeling baseline for aerial data as the most natural next study, and we will release the pretrained backbones so that such a comparison can be reproduced without re-training
\method{}.

\paragraph*{Statistical validation}
All results presented in this paper are based on single-seed pretraining. To partially mitigate the associated concern of run-to-run variability, two deliberate steps were taken. First, every clean
linear-probe number in the paper is produced from two independent linear-head trainings under different random initializations (the linear-eval phase head used in \cref{tab:linear_eval} and the
robustness-phase head used in \cref{tab:robustness}). On the six methods, the absolute difference between the two runs is at most $0.25$ points on all $18$ (method $\times$ dataset) cells, which
places a single-run uncertainty of roughly $\pm 0.2$ on every clean number. Second, the methods that cluster most tightly in the main tables (VICReg, scalar, cosine, and \method{}, all within $0.4$ points on the three-dataset mean in \cref{tab:linear_eval}) cluster the same way in the robustness and OOD tables, which is the behavior one would expect from stable-at-this-scale results. A full three-seed replication remains an important next step. In the present paper, we therefore interpret small differences among VICReg, Cosine gate, and Trust-SSL cautiously and focus the main claim on the large multiplicative vs. additive gap.

\paragraph*{Advantages on Aerial Imagery}
Specifically, with aerial imagery, the additive-residual selective invariance recipe buys three things. First, a consistent $1$--$3$ point improvement in clean linear-probe accuracy relative to SimCLR and BYOL under an identical training budget, which we attribute to the combination of factorized representations, the corruption-family auxiliary classifier, and the additive-residual
alignment. Second, large and reliable gains on information-erasing corruptions at severe levels on EuroSAT. Third, better cross-domain transfer to BDD100K weather splits, visible on
every standard detector and particularly on the hardest (snow) split. Taken together, these findings point to a scoped regime of applicability: the additive-residual selective-invariance recipe is the natural choice when atmospheric or spatial information erasure is the dominant failure mode, whereas in contradiction-heavy or larger-dataset regimes a covariance-regularization baseline such as VICReg remains a strong competitor and should be preferred.

\paragraph*{Current Limitations}
It is important to delineate the limitations of the current formulation. It does not dominate every cell in the robustness table: VICReg beats \method{} on several AID and NWPU cells, and SimCLR is competitive in the weather family across the two larger datasets. The gains compress as dataset complexity grows. The ignorance signal does not behave as classical subjective logic predicts on erasure corruptions. And the $K{+}I$ native OOD score is below Mahalanobis in absolute AUROC terms. We flag these explicitly so that future work can target exactly the right places for improvement.

\section{Conclusion}

This study presented an additive-residual selective invariance framework for self-supervised representation learning in aerial imagery and an evidential instantiation named \method{}. The core contribution is methodological: it is shown that the proper way to incorporate a learned trust signal into an SSL objective is to add a gated stop-gradient residual to the base contrastive loss, rather than to multiply the alignment term by the gate. The highest mean linear-probe accuracy on EuroSAT, AID, and NWPU-RESISC45 among six methods trained under an identical 200-epoch protocol on a 210K aerial corpus is achieved by the proposed \method{} method. The largest improvements are observed on EuroSAT, especially under severe information-erasing corruptions, while competitive but non-leading performance is obtained on AID and NWPU. Moreover, consistent zero-shot transfer to the BDD100K weather splits is achieved, yielding gains of $1$--$3$ Mahalanobis AUROC points over the non-selective baselines. Two principle-testing ablation studies show that the additive-residual formulation is the main source of the observed gain, while the particular choice of trust function is of secondary importance. The evidential instantiation nevertheless retains a distinct value as an interpretable variant, and its use can be extended beyond remote sensing.

\section*{Acknowledgments}
The authors would like to thank Prince Sultan University for its support.

\bibliographystyle{IEEEtran}
\bibliography{main}
\end{document}